\documentclass{article}
\usepackage[preprint]{spconf}
\usepackage{amsmath,amssymb,graphicx,booktabs,xcolor,float,hyperref,url}
\newcommand{\gain}[2]{\textcolor{red}{\scriptsize(+#1)}#2}

\graphicspath{{figures/}}

\title{EXPLORING HETEROGENEOUS MODEL MERGING APPROACH FOR COMPLEX KNOWLEDGE TRANSFER}
\name{Jiahe Fan, Si Chen, Yinghao Hou, Wenbo Xia, Ke Xu, Hong Xie, and Enhong Chen}
\address{%
\begin{tabular}[t]{c}\fontsize{12}{15}\selectfont
University of Science and Technology of China\\
First author email: fanjiahe@mail.ustc.edu.cn
\end{tabular}%
}
\toappear{Preprint.}

\begin{document}
\raggedbottom
\maketitle

\begin{abstract}
Specialized models encode task-oriented behavior, but transferring that behavior to a general language model usually requires training, distillation, or representation alignment.
We study whether such ability can instead be transferred directly at the parameter level.
We apply two existing training-free heterogeneous merging methods, previously shown to transfer knowledge between general language models, to specialist-to-general transfer, projecting a specialist donor into the recipient's shape and interpolating backbone parameters without gradient updates or semantic alignment.
Intersection-Merge (IM) injects a prefix-aligned donor slice matching the recipient shape, while Activate-Prune-Merge (APM) uses forward-pass activation statistics to select which donor dimensions to retain before injection.
Across embedding, reranking, reward modeling, and MoE code-specialist transfer, both methods improve the general recipient, showing that simple heterogeneous merging can move capabilities across diverse specialist roles.
\end{abstract}

\begin{keywords}
heterogeneous model merging, large language models, embedding, reranking, reward modeling, mixture of experts
\end{keywords}

\section{Introduction}
\begin{figure*}[t]
\centering
\includegraphics[width=\textwidth]{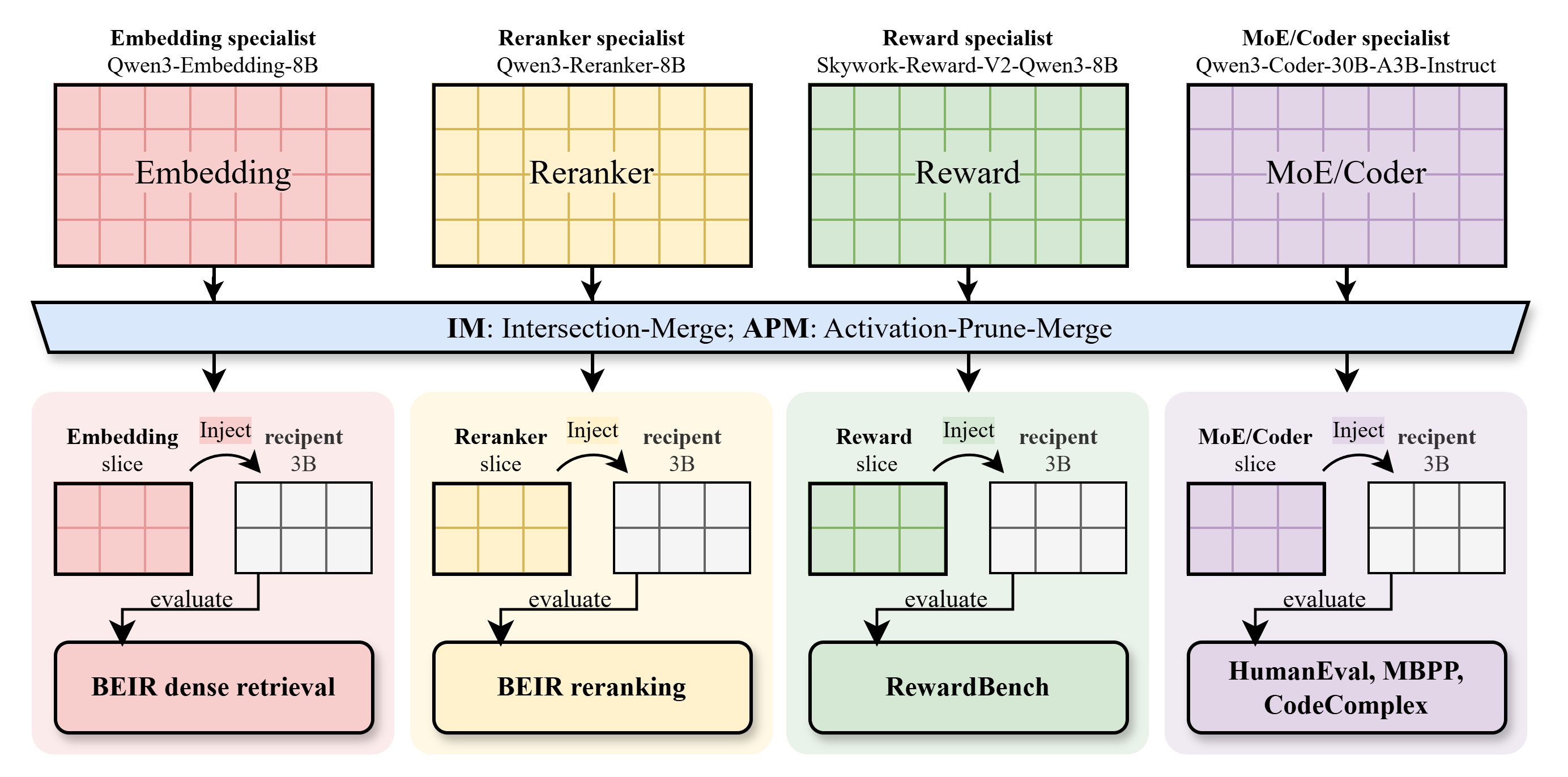}
\caption{Overview of heterogeneous specialist-to-general transfer.
Task-specialized embedding, reranker, reward, and MoE/Coder donors are projected into the common Qwen2.5-3B recipient through the training-free IM/APM parameter-injection framework.
Each transferred model is evaluated on its corresponding task: BEIR reranking, BEIR dense retrieval, RewardBench, or code benchmarks.}
\label{fig:overview}
\end{figure*}

Task-specialized models serve distinct roles: embedding models learn representations for retrieval, rerankers estimate query--document relevance, reward models score response preferences, and code specialists target program synthesis rather than general next-token prediction \cite{reimers2019sbert,karpukhin2020dpr,nogueira2019bertreranking,nogueira2020monot5,liu2025skyworkrewardv2,yang2025qwen3}.
When a specialist is too large for the intended deployment budget, its behavior is usually transferred through supervised fine-tuning or knowledge distillation \cite{hinton2015distilling,wang2020minilm}.
These routes require optimization and, often, examples that connect teacher outputs to student inputs.
Model merging offers a different mechanism: capabilities are manipulated in parameter space, often without additional inference cost \cite{wortsman2022modelsoups,ilharco2023taskarithmetic,yadav2023ties,yu2024dare}.
Many merging methods assume homologous architectures or task models derived from a common initialization.
Heterogeneous merging relaxes this assumption \cite{stoica2023zipit,ainsworth2023gitrebasin} and opens a path for specialist-to-general, large-to-small transfer across distinct model roles.
We ask whether embedding, reranking, reward, and code-specialist donors can improve a general language model on their corresponding tasks through direct heterogeneous parameter injection, without training or semantic alignment.
We test this using Qwen3 embedding and reranking specialists, Skywork-Reward-V2-Qwen3-8B, and the MoE Qwen3-Coder-30B-A3B-Instruct as donors, with Qwen2.5-3B as recipient \cite{yang2025qwen3,qwen2024qwen25,liu2025skyworkrewardv2}.
We study two projection rules.
IM, introduced by Fan et al. \cite{fan2026rethinkingheterogeneousllmmerging}, crops the donor to the recipient's tensor shapes and injects it with a small donor weight.
APM, subsequently introduced by Fan et al. \cite{fan2026trainingfreeknowledgetransfermodel}, extends IM by locating knowledge-carrying donor structures rather than training on activations.
We test this transfer on BEIR, RewardBench, and code benchmarks.
The contributions are threefold. First, we cast specialist-to-general transfer as heterogeneous merging across mismatched functions, not only scale. Second, training-free IM and APM, previously shown between general language models, carry embedding, reranking, reward, and code behavior into one general recipient without semantic alignment. Third, noise controls show that the gains follow donor structure: prefix cropping misses sparsely routed code experts, while activation-guided selection retains them.
Figure~\ref{fig:overview} summarizes the four specialist-to-general transfer settings.

\section{Related Work}
\subsection{Model Merging and Heterogeneous Transfer}
Weight averaging, task arithmetic, interference-aware merging, and parameter dropping combine homologous models without joint training \cite{wortsman2022modelsoups,ilharco2023taskarithmetic,yadav2023ties,yu2024dare}.
A shared architecture leaves these methods with limited guidance once size or computational structure changes.
Distillation and FuseLLM can cross architectures, but only through optimization and data, so they give up training-free merging \cite{hinton2015distilling,wang2020minilm,wan2024fusellm}.
Permutation alignment, ZipIt, and HM3 remain training-free, yet they still assume one architecture, are shown mainly below a billion parameters, and depend on unit or feature alignment \cite{ainsworth2023gitrebasin,jordan2023repair,stoica2023zipit,hackmann2024hm3}.
AdaMMS adds an explicit parameter mapping and coefficient search, and each merged pair stays within one model family \cite{du2025adamms}.
IM and APM avoid both training and semantic alignment, but only for general causal language models that differ in scale \cite{fan2026rethinkingheterogeneousllmmerging,fan2026trainingfreeknowledgetransfermodel}.
Here the donors are embedding, reranker, reward, and sparse MoE-coder specialists, the recipient is fixed as Qwen2.5-3B, and success is measured on each specialist's native task.
\subsection{Pruning and Knowledge Transfer}
One-shot and structured pruning identify parameters or structures that can be removed while retaining behavior \cite{frantar2023sparsegpt,sun2024wanda,ma2023llmpruner}.
Activation-aware criteria motivate using observed internal responses rather than coordinates alone.
Activation-guided pruning further selects donor structures before injection \cite{fan2026trainingfreeknowledgetransfermodel}.
APM uses activation statistics to select coordinates while keeping the recipient frozen and avoiding representation matching.
\subsection{Specialist Model Roles}
Dense encoders learn vector spaces for efficient retrieval \cite{reimers2019sbert,karpukhin2020dpr,izacard2022contriever,wang2022e5,muennighoff2022sgpt}, while rankers score query--document relevance directly \cite{nogueira2019bertreranking,nogueira2020monot5}.
Reward models instead score response preferences, and code specialists target program synthesis.
Qwen3 supplies related embedding and reranking specialists \cite{zhang2025qwen3embedding}.
We study whether task bias from these distinct roles survives heterogeneous projection, including projection from a sparse MoE code donor into a dense recipient.

\section{Method}
Let a specialist donor have parameters $\theta_D$ and a general recipient have parameters $\theta_R$.
For a projection rule $P$ and injection ratio $\alpha$, IM and APM inject projected backbone tensors as
\begin{equation}
\theta'_R(k)=
\begin{cases}
(1-\alpha)\theta_R(k)+\alpha P(\theta_D)(k), & k\in\mathcal{K}_P,\\
\theta_R(k), & \text{otherwise},
\end{cases}
\label{eq:merge}
\end{equation}
where $\mathcal{K}_P$ denotes projected Transformer-backbone tensors; token embeddings and the language-model head are kept from the recipient.
\subsection{Intersection-Merge}
IM provides the coordinate-based baseline.
For every selected donor tensor and its recipient counterpart, it copies the prefix intersection along each dimension into a recipient-shaped shell.
Layers outside the recipient depth and coordinates outside the recipient width are discarded.
The projected layer, attention, MLP, and normalization tensors are then injected using Eq.~\eqref{eq:merge}.
This deterministic operation requires neither calibration data nor gradients, but its coordinate choice is agnostic to donor activity.
On every axis the retained block is the leading index range, so recipient layer $\ell$ and channel $j$ take donor coordinates $(\ell,j)$ when both exist.
No permutation is applied, so a specialist coordinate that lies outside this prefix is never injected.
\subsection{Activation-Guided Pruning}
APM runs the frozen donor on calibration sequences and ranks coordinates by mean absolute activation.
Hidden channels are selected once, and MLP neurons separately in each retained layer:
\begin{equation}
\mathcal{I}_{d}=\operatorname{TopK}(S_{\mathrm{hidden}},d_{s}),\quad
\mathcal{I}_{m}^{(\ell)}=\operatorname{TopK}(S_{\mathrm{mlp}}^{(\ell)},m_{s}),
\label{eq:topk}
\end{equation}
where $d_{s}$ and $m_{s}$ match the recipient widths.
Query and key--value heads are chosen the same way, summing scores inside each grouped-query group, and every index set is sorted to preserve donor order.
With the retained layers, these indices form $\mathcal{S}_{D}$ and extract a recipient-shaped slice for Eq.~\eqref{eq:merge}:
\begin{equation}
P(\theta_{D})=\mathcal{P}_{D}(\theta_{D};\mathcal{S}_{D},\mathcal{A}_{s}),
\label{eq:apm}
\end{equation}
with $\mathcal{A}_{s}$ the recipient shape.
Coupled axes stay together, and the activations never enter a parameter update.
\subsection{Specialist-to-General Transfer}
The projection-and-injection framework is evaluated across four specialist roles.
The reranking and embedding lines use Qwen3-Reranker-8B and Qwen3-Embedding-8B and retain their retrieval evaluations.
The reward line uses Skywork-Reward-V2-Qwen3-8B and evaluates pairwise preference scoring.
The code line transfers from the sparse MoE Qwen3-Coder-30B-A3B-Instruct and evaluates code generation.
\section{Experimental Setup}
\label{sec:setup}
\subsection{Models}
The recipient is the 3.09B-parameter, 36-layer Qwen2.5-3B \cite{qwen2024qwen25}.
Retrieval donors are Qwen3-Reranker-8B and Qwen3-Embedding-8B \cite{zhang2025qwen3embedding,yang2025qwen3}; the other donors are Skywork-Reward-V2-Qwen3-8B \cite{liu2025skyworkrewardv2} and the sparse MoE Qwen3-Coder-30B-A3B-Instruct \cite{yang2025qwen3}.
\subsection{Benchmarks}
We evaluate retrieval on 11 BEIR datasets \cite{thakur2021beir}: ArguAna, CQADupStack, DBPedia-Entity, FiQA, HotpotQA, NFCorpus, NQ, Quora, SciFact, TREC-COVID, and Touche-2020.
\subsection{Evaluation Protocol}
Retrieval uses NDCG@10. Reranking scores TF--IDF candidates via yes/no log probabilities; embedding applies weighted-mean pooling with L2 normalization. RewardBench reports category accuracies and order diagnostics \cite{lambert2025}; on code tasks we report accuracy on HumanEval \cite{liu2023evalplus}, MBPP \cite{austin2021mbpp}, and CodeComplex \cite{baik2025codecomplex}.
APM activation examples are used only to locate knowledge-carrying coordinates, never to train, and are strictly disjoint from the test sets.
\subsection{IM/APM Search Spaces}
We sweep $\alpha\in\{0.005,0.010,0.015,0.020,0.025\}$ for IM, APM, and Noise.
Main tables report one locked checkpoint (Avg.). Env.\ is the envelope mean; the reranker per-dataset envelope is in Table~\ref{tab:reranker-main}(b). RewardBench is a single set.
Full per-dataset and per-benchmark envelopes are reported in Appendix~\ref{sec:oracle}.
\section{Results}
\subsection{Reranker Results}
\begin{table}[t]
\centering
\small
\setlength{\tabcolsep}{2.8pt}
\caption{Reranker NDCG@10 (\%). Donor: Qwen3-Reranker-8B.
(a)~Locked. (b)~Per-dataset oracle envelope.}
\label{tab:reranker-main}
\begin{tabular*}{\columnwidth}{@{\extracolsep{\fill}}lrrrr}
\multicolumn{5}{l}{\textbf{(a) Locked checkpoint}} \\
\toprule
Dataset & 3B & IM & APM & Noise \\
\midrule
ArguAna & 7.14 & \gain{3}{\textbf{10.08}} & 7.06 & 7.24 \\
CQADupStack & 2.92 & 2.86 & \textbf{3.19} & 2.70 \\
DBPedia-Entity & 12.90 & \gain{1}{\textbf{14.30}} & 13.66 & 12.75 \\
FiQA & 4.80 & \textbf{5.62} & 5.06 & 4.60 \\
HotpotQA & 8.93 & \gain{6}{\textbf{14.89}} & 9.24 & 7.63 \\
NFCorpus & 8.40 & \gain{2}{\textbf{10.22}} & 8.44 & 10.08 \\
NQ & 2.78 & \gain{8}{\textbf{10.41}} & 3.70 & 2.79 \\
Quora & \textbf{41.49} & 27.13 & 41.01 & 40.51 \\
SciFact & 23.15 & 19.52 & \gain{2}{\textbf{25.49}} & 23.28 \\
TREC-COVID & 27.62 & \gain{17}{\textbf{44.95}} & \gain{11}{39.10} & 36.10 \\
Touche-2020 & 11.10 & 4.20 & 10.94 & \textbf{12.05} \\
\midrule
Avg. & 13.75 & \gain{1}{14.92} & \gain{1}{\textbf{15.17}} & 14.52 \\
\bottomrule
\end{tabular*}

\vspace{6pt}

\begin{tabular*}{\columnwidth}{@{\extracolsep{\fill}}lrrrr}
\multicolumn{5}{l}{\textbf{(b) Oracle envelope}} \\
\toprule
Dataset & 3B & IM & APM & Noise \\
\midrule
ArguAna & 7.14 & \gain{6}{\textbf{13.11}} & \gain{6}{12.87} & 7.68 \\
CQADupStack & 2.92 & 3.17 & \textbf{3.56} & 2.87 \\
DBPedia-Entity & 12.90 & \gain{1}{\textbf{14.30}} & 13.77 & 12.75 \\
FiQA & 4.80 & \textbf{5.62} & 5.06 & 4.96 \\
HotpotQA & 8.93 & \gain{6}{14.89} & \gain{10}{\textbf{18.86}} & 8.33 \\
NFCorpus & 8.40 & \gain{2}{10.24} & 9.26 & \textbf{10.40} \\
NQ & 2.78 & \gain{8}{\textbf{10.41}} & \gain{7}{9.64} & 2.94 \\
Quora & 41.49 & \gain{2}{\textbf{43.88}} & \gain{2}{43.34} & 41.74 \\
SciFact & 23.15 & \gain{4}{26.84} & \gain{5}{\textbf{28.35}} & 23.64 \\
TREC-COVID & 27.62 & \gain{17}{44.95} & \gain{23}{\textbf{50.86}} & 38.29 \\
Touche-2020 & 11.10 & 10.54 & 11.50 & \textbf{13.95} \\
\midrule
Avg. & 13.75 & \gain{4}{18.00} & \gain{5}{\textbf{18.82}} & 15.23 \\
\bottomrule
\end{tabular*}
\end{table}

Average NDCG@10 rises from 13.75\% to 14.92\% (IM) and 15.17\% (APM).
The largest gains are TREC-COVID (27.62 to 44.95 / 39.10), NQ (2.78 to 10.41, IM), and HotpotQA (8.93 to 14.89, IM).
The envelope lifts the mean to 18.00\% and 18.82\%, and TREC-COVID to 50.86 under APM.
\subsection{Embedding Results}
\begin{table}[t]
\centering
\small
\setlength{\tabcolsep}{3.5pt}
\caption{Embedding NDCG@10 (\%). Donor: Qwen3-Embedding-8B. Avg.: locked mean. Env.: envelope mean.}
\label{tab:embedding-main}
\begin{tabular*}{\columnwidth}{@{\extracolsep{\fill}}lrrrr}
\toprule
Dataset & 3B & IM & APM & Noise \\
\midrule
ArguAna & 21.82 & 22.71 & \gain{1}{\textbf{23.00}} & 22.38 \\
CQADupStack & 4.24 & 4.36 & \textbf{4.61} & 4.25 \\
DBPedia-Entity & 0.57 & 0.69 & \textbf{0.82} & 0.62 \\
FiQA & 4.11 & 3.61 & 3.76 & \textbf{4.12} \\
HotpotQA & 2.62 & \textbf{2.94} & 2.80 & 2.65 \\
NFCorpus & \textbf{1.91} & 1.63 & 1.78 & 1.90 \\
NQ & \textbf{0.42} & 0.41 & 0.36 & 0.40 \\
Quora & 54.51 & 54.42 & \textbf{55.36} & 54.69 \\
SciFact & 23.28 & \gain{2}{25.60} & \gain{3}{\textbf{26.64}} & 24.24 \\
TREC-COVID & 12.63 & \gain{2}{\textbf{14.20}} & 13.30 & 12.62 \\
Touche-2020 & 1.63 & \textbf{2.10} & 1.85 & 1.69 \\
\midrule
Avg. & 11.61 & 12.06 & \textbf{12.21} & 11.78 \\
Env. & 11.61 & 12.22 & \gain{1}{\textbf{12.77}} & 11.95 \\
\bottomrule
\end{tabular*}
\end{table}

Average NDCG@10 rises from 11.61\% to 12.06\% (IM) and 12.21\% (APM), led by SciFact (23.28 to 26.64, APM) and TREC-COVID (12.63 to 14.20, IM).
\subsection{Reward and MoE/Coder Results}
\begin{table}[t]
\centering
\small
\setlength{\tabcolsep}{3.5pt}
\caption{Reward-model results (\%). Donor: Skywork-Reward-V2-Qwen3-8B; 3B recipient. RewardBench is a single benchmark.}
\label{tab:reward-results}
\begin{tabular*}{\columnwidth}{@{\extracolsep{\fill}}lrrrr}
\multicolumn{5}{l}{\textbf{(a) Performance metrics}} \\
\toprule
Metric & 3B & IM & APM & Noise \\
\midrule
Overall & 76.83 & \textbf{77.41} & \textbf{77.41} & 76.25 \\
Chat & 63.80 & \gain{5}{68.71} & \gain{2}{66.26} & \textbf{69.33} \\
Chat Hard & 52.77 & \gain{2}{\textbf{54.55}} & \gain{1}{53.88} & 52.11 \\
Reasoning & 89.78 & 89.48 & \textbf{90.38} & 88.66 \\
Safety & 70.16 & \textbf{70.62} & 69.84 & 69.22 \\
\bottomrule
\end{tabular*}

\vspace{4pt}

\begin{tabular*}{\columnwidth}{@{\extracolsep{\fill}}lrrrr}
\multicolumn{5}{l}{\textbf{(b) Order-related metrics}} \\
\toprule
Metric & 3B & IM & APM & Noise \\
\midrule
Original order & 60.00 & 57.37 & \textbf{60.12} & 50.17 \\
Swapped order & 64.10 & \gain{3}{66.85} & 63.83 & \textbf{72.88} \\
Order gap & 4.10 & 9.48 & \textbf{3.71} & 22.71 \\
Order consistency & 33.58 & \textbf{33.69} & 33.58 & 30.72 \\
\bottomrule
\end{tabular*}
\end{table}

\begin{table}[t]
\centering
\small
\setlength{\tabcolsep}{2pt}
\caption{MoE/Coder results (\%). Donor: Qwen3-Coder-30B-A3B-Instruct. Avg.: locked mean. Env.: envelope mean.}
\label{tab:moe-results}
\begin{tabular*}{\columnwidth}{@{\extracolsep{\fill}}lrrrr}
\toprule
Benchmark & 3B & IM & APM & Noise \\
\midrule
HumanEval & 80.49 & 80.49 & \gain{1}{\textbf{81.71}} & 78.05 \\
MBPP & 59.21 & 59.65 & \textbf{60.09} & 59.65 \\
CodeComplex & 36.12 & 33.73 & \gain{3}{\textbf{38.76}} & 34.93 \\
\midrule
Avg. & 58.61 & 57.96 & \gain{2}{\textbf{60.18}} & 57.54 \\
Env. & 58.61 & 58.75 & \gain{3}{\textbf{61.12}} & 58.15 \\
\bottomrule
\end{tabular*}
\end{table}

On RewardBench, Overall rises from 76.83\% to 77.41\% (Table~\ref{tab:reward-results}), with the larger category gain on Chat (63.80 to 68.71, IM).
APM also keeps a smaller order gap (3.71\% vs.\ 9.48\%).
On MoE/Coder transfer, APM raises the mean from 58.61\% to 60.18\%, led by CodeComplex (36.12 to 38.76) and HumanEval (80.49 to 81.71).
\subsection{Ablation Against Gaussian Noise}
To test whether improvements arise from donor structure rather than perturbation alone, we inject Gaussian noise into the same recipient tensors using the same five ratios. Noise remains below locked APM on average retrieval: 14.52\% versus 15.17\% for reranking and 11.78\% versus 12.21\% for embedding. In RewardBench, noise reaches 76.25\% Overall, below IM and APM (both 77.41\%), with a larger order gap (22.71\%). The code control is also below APM on HumanEval (78.05\% versus 81.71\%), MBPP (59.65\% versus 60.09\%), and CodeComplex (34.93\% versus 38.76\%). These results suggest that the strongest gains are not explained by arbitrary perturbations alone, although noise can occasionally help on individual datasets.

\section{Analysis}
Related Qwen backbones appear to retain compatible low-level structure, so small $\alpha$ can inject specialist bias without overwriting the recipient.
IM and APM are simple heterogeneous merging strategies rather than one shared checkpoint: the preferred $\alpha$ varies across tasks and datasets, a seesaw effect that is well documented for linear merging \cite{ilharco2023taskarithmetic}.
A five-point $\alpha$ grid is therefore enough to expose their potential; the embedding and MoE envelope means are reported in Tables~\ref{tab:embedding-main} and~\ref{tab:moe-results}.
APM's activation-guided selection helps reranking, embedding, and code, while IM remains competitive for reward and some reranking settings.
Locked IM on the MoE coder is slightly below the recipient (57.96 vs.\ 58.61), because prefix cropping misses routed experts while APM retains them.

\section{Conclusion}
IM and APM enable training-free specialist transfer without semantic alignment.
Across BEIR, reward, and code tasks, APM improves the recipient; IM improves retrieval and reward, matches APM on reward Overall, and is complementary on reranking.
These results support training-free transfer across specialist roles.

\section*{Acknowledgment}
Supported by the CAS Strategic Priority Research Program (LLM Mutual Learning Mechanisms and Methods).
This support enabled the study of training-free transfer from specialist donors into one general language model.
We thank the maintainers of the public models and benchmarks used in the experiments.

\newcounter{savedsection}
\setcounter{savedsection}{\value{section}}
\appendix
\section{Oracle Envelopes}
\label{sec:oracle}
Each IM, APM, and Noise score below is the best value on the same five-ratio grid $\alpha\in\{0.005,0.010,0.015,0.020,0.025\}$ used in the main paper. APM further varies activation position. These tables are not a single deployable checkpoint.

\begin{table}[H]
\centering
\small
\setlength{\tabcolsep}{2.8pt}
\caption{Reranker NDCG@10 (\%). Donor: Qwen3-Reranker-8B. Per-dataset best $\alpha$ on a 5-point grid (oracle envelope).}
\label{tab:reranker-oracle}
\begin{tabular*}{\columnwidth}{@{\extracolsep{\fill}}lrrrr}
\toprule
Dataset & 3B & IM & APM & Noise \\
\midrule
ArguAna & 7.14 & \gain{6}{\textbf{13.11}} & \gain{6}{12.87} & 7.68 \\
CQADupStack & 2.92 & 3.17 & \textbf{3.56} & 2.87 \\
DBPedia-Entity & 12.90 & \gain{1}{\textbf{14.30}} & 13.77 & 12.75 \\
FiQA & 4.80 & \textbf{5.62} & 5.06 & 4.96 \\
HotpotQA & 8.93 & \gain{6}{14.89} & \gain{10}{\textbf{18.86}} & 8.33 \\
NFCorpus & 8.40 & \gain{2}{10.24} & 9.26 & \textbf{10.40} \\
NQ & 2.78 & \gain{8}{\textbf{10.41}} & \gain{7}{9.64} & 2.94 \\
Quora & 41.49 & \gain{2}{\textbf{43.88}} & \gain{2}{43.34} & 41.74 \\
SciFact & 23.15 & \gain{4}{26.84} & \gain{5}{\textbf{28.35}} & 23.64 \\
TREC-COVID & 27.62 & \gain{17}{44.95} & \gain{23}{\textbf{50.86}} & 38.29 \\
Touche-2020 & 11.10 & 10.54 & 11.50 & \textbf{13.95} \\
\midrule
Avg. & 13.75 & \gain{4}{18.00} & \gain{5}{\textbf{18.82}} & 15.23 \\
\bottomrule
\end{tabular*}
\end{table}

\begin{table}[H]
\centering
\small
\setlength{\tabcolsep}{3.5pt}
\caption{Embedding NDCG@10 (\%). Donor: Qwen3-Embedding-8B. Per-dataset best $\alpha$ on a 5-point grid (oracle envelope).}
\label{tab:embedding-oracle}
\begin{tabular*}{\columnwidth}{@{\extracolsep{\fill}}lrrrr}
\toprule
Dataset & 3B & IM & APM & Noise \\
\midrule
ArguAna & 21.82 & \gain{1}{23.05} & \gain{2}{\textbf{23.33}} & 22.38 \\
CQADupStack & 4.24 & 4.36 & \textbf{4.74} & 4.31 \\
DBPedia-Entity & 0.57 & 0.69 & \textbf{0.96} & 0.62 \\
FiQA & 4.11 & 4.22 & \textbf{4.57} & 4.24 \\
HotpotQA & 2.62 & 2.94 & \textbf{3.07} & 2.65 \\
NFCorpus & 1.91 & 2.03 & \gain{1}{\textbf{3.22}} & 1.95 \\
NQ & 0.42 & 0.47 & \textbf{0.51} & 0.40 \\
Quora & 54.51 & 54.66 & \gain{1}{\textbf{55.53}} & 54.69 \\
SciFact & 23.28 & \gain{2}{25.60} & \gain{3}{\textbf{26.64}} & 24.24 \\
TREC-COVID & 12.63 & \gain{2}{14.20} & \gain{3}{\textbf{15.62}} & 14.05 \\
Touche-2020 & 1.63 & 2.20 & \textbf{2.29} & 1.91 \\
\midrule
Avg. & 11.61 & 12.22 & \gain{1}{\textbf{12.77}} & 11.95 \\
\bottomrule
\end{tabular*}
\end{table}

\begin{table}[H]
\centering
\small
\setlength{\tabcolsep}{2pt}
\caption{MoE/Coder results (\%). Donor: Qwen3-Coder-30B-A3B-Instruct. Per-benchmark best $\alpha$ (oracle envelope).}
\label{tab:moe-oracle}
\begin{tabular*}{\columnwidth}{@{\extracolsep{\fill}}lrrrr}
\toprule
Benchmark & 3B & IM & APM & Noise \\
\midrule
HumanEval & 80.49 & 80.49 & \gain{2}{\textbf{82.93}} & 79.88 \\
MBPP & 59.21 & 59.65 & \gain{2}{\textbf{60.96}} & 59.65 \\
CodeComplex & 36.12 & 36.12 & \gain{3}{\textbf{39.47}} & 34.93 \\
\midrule
Avg. & 58.61 & 58.75 & \gain{3}{\textbf{61.12}} & 58.15 \\
\bottomrule
\end{tabular*}
\end{table}

\setcounter{section}{\value{savedsection}}
\renewcommand{\thesection}{\arabic{section}}
\bibliographystyle{IEEEbib}
\bibliography{references}

\end{document}